\documentclass[11pt,a4paper]{article}
\usepackage[hyperref]{emnlp-ijcnlp-2019}
\usepackage{times}
\usepackage{latexsym}
\usepackage{tabularx}
\usepackage{amsmath}
\usepackage[ruled,longend]{algorithm2e}

\usepackage{url}

\aclfinalcopy 

\usepackage{graphicx}
\usepackage{multirow}
\usepackage{graphicx}

\title{BIG MOOD: Relating Transformers to Explicit Commonsense Knowledge}

\author{Jeff Da \\
   Paul G.~Allen School of Computer Science \& Engineering,\\University of Washington, Seattle, WA, USA \\
  {\tt jzda@cs.washington.edu} \\}

\date{}

\begin{document}
\maketitle
\begin{abstract}
We introduce a simple yet effective method of integrating contextual embeddings with commonsense graph embeddings, dubbed BERT Infused Graphs: Matching Over Other embeDdings. First, we introduce a preprocessing method to improve the speed of querying knowledge bases. Then, we develop a method of creating knowledge embeddings from each knowledge base. We introduce a method of aligning tokens between two misaligned tokenization methods. Finally, we contribute a method of contextualizing BERT after combining with knowledge base embeddings. We also show BERTs tendency to correct lower accuracy question types. Our model achieves a higher accuracy than BERT, and we score fifth on the official leaderboard of the shared task and score the highest without any additional language model pretraining.
\end{abstract}

\section{Introduction}

Recently, wide-scale pre-training and deep contextual representations have taken the world by storm. \citet{Peters2018DeepCW} underscored the importance of bidirectional contextual representations by using traditional neural networks trained on a large corpus of text. \citet{Devlin2018BERTPO} used transformers \cite{Vaswani2017AttentionIA} and word masking to pre-train on another large corpus of data, reporting human-level performance on one commonsense dataset \cite{Zellers2018SWAGAL}. \citet{Yang2019XLNetGA} achieves state-of-the-art on RACE \cite{Lai2017RACELR} with a Transformer-XL based model \cite{Dai2019TransformerXLAL}.

The key to success in the performance of many of these models is their ability to train on extremely large datasets. BERT \cite{Devlin2018BERTPO}, for example, trains on the BooksCorpus \cite{Zhu2015AligningBA} and English Wikipedia, for a combined 3,200M words. Other iterations increased the amount of knowledge used during pre-training, such as RoBERTa \cite{Liu2019RoBERTaAR}. Training large-scale models on these massive datasets has drawbacks, such as significantly increased carbon pollution and harm to the environment \cite{Schwartz2019GreenA, Strubell2019EnergyAP}.

We present a methodology of combining queries from commonsense knowledge bases with contextual embeddings, \textbf{BIG MOOD} - BERT Infused Graphs: Matching Over Other embeDdings, and abbreviated for its relationship to human knowledge awareness. Our methodology achieves a increase without significant additional fine-tuning or pre-training. Instead, it learns a separate representation from commonsense graphical knowledge bases, and augments the BERT representation with this learned explicit representation. We introduce several methods of combining and querying knowledge base embeddings to introduce them to the BERT embedding layers.

\section{Related Work}

\subsection{Knowledge Graphs}

Significant research has been put into representing human knowledge in various ways \cite{Lenat1989BuildingLK, Auer2007DBpediaAN, Chambers2008UnsupervisedLO}. ConceptNet \cite{Speer2013ConceptNet5A} contains various aspects of commonsense knowledge through a knowledge graph.The knowledge is collected from crowed-sourced resources \cite{Meyer2012WiktionaryAN, Havasi2010OpenMC, Ahn2006VerbosityAG} and expert-created resources \cite{Miller1992WORDNETAL, Breen2004JMdictAJ}. WebChild \cite{tandon-etal-2017-webchild} is a collection of commonsense knowledge automatically extracted from web contents. The database is constructed similarly to ConceptNet, and intended to cover concepts that ConceptNet does not cover. ATOMIC \cite{Sap2018ATOMICAA} focuses on inferential $If-Then$ relations, built for everyday commonsense reasoning.

\subsection{Knowledge Integration}

Knowledge graphs have been applied in various natural language processing applications, such as reading comprehension \cite{Lin2017ReasoningWH, Yang2017LeveragingKB} and machine translation \cite{Zhang2017PriorKI}. ERNIE: Enhanced Representation through Knowledge Integration \cite{Sun2019ERNIEER} appends knowledge to the input of the model and learns via knowledge masking, as well as entity-level masking and phrase-level masking. TriAN \cite{Wang2018YuanfudaoAS}, the top public model on the MCScript \cite{Ostermann2018MCScriptAN} shared task, uses ConceptNet embeddings to highlight relationships between the question, text, and answer.

\section{Model}

\begin{table}[]
\begin{tabularx}{0.46\textwidth}{X} \hline
Passage: I had decided that I wanted to visit my friend Paul whom lives quite a distance away. With this and my fear of air travel in mind I decided to take a train. After researching and finding one online I was well on my way to going to see my friend Paul. I drive to the station and decide that I am going to purchase a round trip ticket as this would be cheaper than just buying both tickets separately. Whenever my train arrives I have to get in line as they process our tickets. After all this is done I decide to take a seat by the window. I sit and fall asleep a bit as I ride on the train for hours. After a couple hours we finally reach the destination and I get off the train, excited to see my friend. \\ \hline
When did they wait for their train? \\
a) before buying the ticket  \\
\textbf{b) after buying a ticket} \\ \hline
\end{tabularx}
\caption{Example of a prompt from the shared task dataset, an everyday commonsense reasoning dataset. Questions often require script knowledge that extends beyond referencing the text.}
\label{examplemc}
\end{table}

\begin{figure*} 
    \centering
  \includegraphics[width=\textwidth]{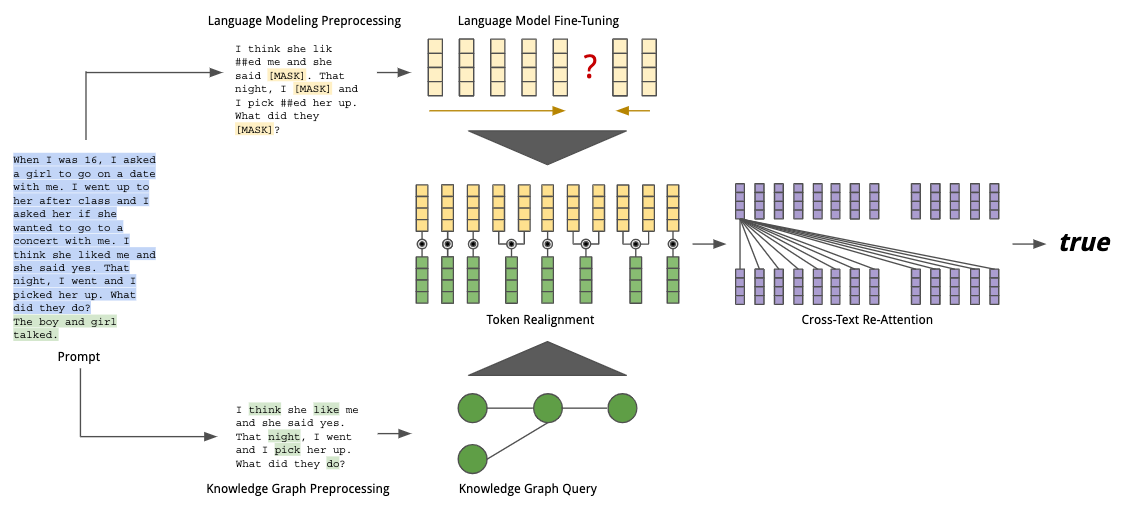}
\caption{Our model architecture. Our design mimics \cite{Vaswani2017AttentionIA}. Since the queries work on whole words only, one knowledge base embeddings may be integrated with one or more language embedding. Several self-attention encoding layers are used.}
\label{fig:my_label}
\end{figure*}

We present our model for this shared task. Our model has three major components: language model adaptation, knowledge graph embeddings, and attention for classification.

\subsection{Data Preprocessing}

Before model usage, we preprocess the data in two ways to make it easier for the model to understand. For language modeling, we create training data similar to those in BERT \cite{Devlin2018BERTPO}. For knowledge graph use, we preprocess language to create commonsense object and relationship vocabulary and to match as many related commonsense objects as possible.

\subsubsection{Language Model Preprocessing}

We prepossess each passage for training. We use this process for each training epoch, since it allows for the most dense pretraining framework.

Commonly known as a \textit{cloze task}, \citet{Devlin2018BERTPO} introduced a framework that pretrained transformers \cite{Vaswani2017AttentionIA} based on masked token prediction. First, we preprocess the tokens with WordPiece embeddings \cite{Wu2016GooglesNM}. Then, we append special $[CLS]$ and $[SEP]$ to each datum. We append $[CLS]$ to the beginning of each datum, and $[SEP]$ to separate the question with the answer, as such:
$$
[CLS]\ passage\ question\ [SEP]\ ans.\ [SEP]
$$

Then, we randomly mask 15\% of all WordPiece embeddings. Unlike \citet{Devlin2018BERTPO}, we run the randomization script once per each training epoch. Otherwise, we follow the procedure in \citet{Devlin2018BERTPO}. 80\% of the time, we replace the word with the $[MASK]$ prediction, to be replaced through cloze task prediction. 10\% of the time, we replace the word with a random word. 10\% of the time, we keep the word unchanged.

Combined with the above cloze task, we process the data for next sentence prediction. We do this process after the cloze task masking, similar to \citet{Devlin2018BERTPO}. For each datum, we randomly pick either a sentence labeled correctly as the next sentence 50\% of the time, or a random sentence 50\% of the time. We ensure that the random sentence is not the next sentence.

\subsubsection{Knowledge Graph Processing}

\begin{algorithm}
  \SetAlgoLined
  \DontPrintSemicolon
  \For{prompt in dataset}{
    \For{KG in knowledge\_graphs}{
        \For{(start, end, edge) in KG}{
                \If{start in prompt \& end in prompt}{
                    add((start, end, edge))\;
                    index\_as\_relationship(edge)
                }
            }
        }
    }
        
  \caption{Knowledge Graph Vocab Creation}
\end{algorithm}

We preprocess the data in the shared task along with knowledge graph preprocessing. The purpose of this procedure is to reduce the number of items in the knowledge graph, to speed up fine-tuning since the knowledge graphs are extremely large, and also to ensure matching between as many different types of knowledge graph edges that are relevant as possible.

First, we create an index of \textit{(start, end, edge)} relationships that match vocabulary within the shared task prompt. For each \textit{(start, end, edge)}, we check to see if there are any matching prompts in which \textit{start} is present in the text and \textit{end} is present in the text. If so, we store the \textit{(start, end, edge)}, and note the edge as a relationship. We also index the relationship (edge), giving an index for each unique relationship.

For longer sequences, we allow matches between any trigram, and store an index for each trigram matched. In addition, we stem words beforehand, to ensure that the different word endings do not effect the result of the matches. We use the Porter Stemmer \cite{Porter1980AnAF} to stem each word in both the text and the knowledge graph. Note that we only use the stemming to match different words, and do not keep the stemmed words for later use in the process, as to keep comparability between embedding types. We also stem words in knowledge bases, to allow for comprasion. Algorithm 1 shows our process for matching sequences.

\subsection{Knowledge Graph Usage}

We query each of three knowledge bases to create an embedding layer, for each word, for each knowledge graph. Here, we describe our procedure for querying each knowledge graph. We stem words beforehand, to allow for matches agnostic of linguistic postfixes \cite{Merkhofer2018MITREAS}.

\subsubsection{ConceptNet}

ConceptNet \cite{Speer2013ConceptNet5A} represents everyday words and phrases, with edges between the commonsense relationships between them. We first preprocess ConceptNet, keeping only the vocabulary present in the shared task. Then, for each edge, we store a tuple $(agent,\ dependent,\ relationship)$ that describes the commonsense relationship mentioned in the knowledge graph.

During fine-tuning, we check the text for any present $agent$, $dependent$ pairs. If any word in the text is an $agent$, and the $dependent$ is present in the text, we add that $relationship$ index as input into the embedding layer. (For $agent$s that span more than one word, such as the phrase "apple pie", we apply the index to the first word, as long as the entire phrase is found in the text). We randomly generate a length 10 embedding for each relationship, and if more than one relationship is matched, we randomly pick one.

\subsubsection{WebChild}

WebChild \cite{tandon-etal-2017-webchild} is a large collection of commonsense knowledge collected from various sources on the web. The format is similar to ConceptNet, which allows us to follow a similar process. WordNet instances are split into categories $part-whole$, $comparative$, $property$, $activity$, and $spatial$. For each category, we capture the $(agent,\ dependent,\ relationship)$ tuple, which is usually defined as properties such as $x_{disambi}$, $y_{disambi}$, and $sub-relation$, but is slightly different for each category. We ignore the WordNet \cite{Miller1992WORDNETAL} relation (some categories will contain subjects such as $bike\#n\#1$, and take only the stemmed word. For fine-tuning, we follow the same procedure as ConceptNet, creating an additional 10-length embedding for each word.

\subsubsection{ATOMIC}

ATOMIC \cite{Sap2018ATOMICAA} is a resource that focuses on inferential knowledge via $If-Then$ relations. ATOMIC separates its relationships into nine different types ($xNeed$, $xIntent$, $xAttr$, $xEffect$, $xReact$, $xWant$, $oEffect$, $oWant$). For each of the nine categories, for each datum in the given category, we search our text for relationships that match the defined $If-Then$ relationship. Since each relationship is nearly a full sentence, we allow a match to be any trigram matched between the given datum and the text. Then, we append an index $[0, 8]$ to the embedding layer of the first word in the selected trigram based on the type of relationship matched. For fine-tuning, we follow the same procedure as ConceptNet and WebChild, creating an additional 10-length embedding for each word. 

\subsection{Architecture}

Out modeling procedure consists of three parts. First, we query each knowledge graph, allowing us to create embeddings for each specific graph. Then, we describe our word-level knowledge fusion procedure, creating augmented embeddings for each word. Finally, we describe our fine-tuning procedure for the shared task dataset. We modify pytorch-transformers\footnote{https://github.com/huggingface/pytorch-transformers}.

\subsubsection{Language Model Fine-Tuning}

Contrary to \citet{Devlin2018BERTPO}, we do language model fine-tuning in addition to classification fine-tuning. We find that this generally provides better results, and allows for more stable accuracy since the shared task involves a small dataset. For each prompt, we use the previous preprocessed data to create tasks for our model to predict. We do this before token realignment, so this happens before any extra knowledge graph embeddings are added to the model architecture. For masked tokens, we predict that token through bidirectional context, the same as \citet{Devlin2018BERTPO}. For next sentence prediction, we use the unbiased method previously introduced as well as in \citet{Devlin2018BERTPO}. 

\subsubsection{Token Realignment}

\begin{algorithm*}
  \SetAlgoLined
  \DontPrintSemicolon
  \textbf{token\_realignment(seq\_1, seq\_2):}\;
  alignment\_dict = dict\;
  seq\_1\_i = 0\;
  seq\_2\_i = 0\;
  \While{seq\_1\_i \textless len(seq\_1) \& seq\_2\_i \textless len(seq\_2)}{
    \If{seq\_1[seq\_1\_i] is seq\_2[seq\_2\_i]}{
        alignment\_dict[seq\_1\_i].append(seq\_2\_i)\;
        seq\_1\_i++\;
        seq\_2\_i++\;
    }
    \If{seq\_1[seq\_1\_i] in seq\_2[seq\_2\_i]}{
        alignment\_dict[seq\_1\_i].append(seq\_2\_i)\;
        seq\_1\_i++\;
    }
    \If{seq\_2[seq\_2\_i] in seq\_1[seq\_1\_i]}{
        alignment\_dict[seq\_1\_i].append(seq\_2\_i)\;
        seq\_2\_i++\;
    }
  }
  \textbf{return} alignment\_dict
  \caption{Psuedocode for the token realignment algorithm, a method of finding token alignments between two different sequences.}
\end{algorithm*}

We do a word-level fusion to incorporate knowledge embeddings into the BERT model. First, we collect word embeddings from BERT. We sum the last four layer of BERT together, as suggested by "The Illustrated BERT, ELMo, and co." \footnote{http://jalammar.github.io/illustrated-bert/}. We fuse these embeddings with the embeddings gathered from querying each of the three databases. For each word, we take the dyadic product, or linear fusion, of the contextual BERT embeddings with the concatenation of the three graph embeddings. When there is no related embedding (if the word did not match any edges during querying, or if the word is a BERT-specific token such as $[CLS]$, we do not do any dyadic fusion. Finally, to get a single linear layer, we concatenate each dimension of the result of the dyadic fusion with the original BERT embedding. Algorithm 2 shows a detailed explanation of our token realignment process.

\subsubsection{Re-Attention}

To get a final result, we do a few more necessary steps. First, we do a single layer of self-attention over the text, allowing each of the word-level embeddings to interact with one another. For this self-attention, we follow the same process as in \cite{Vaswani2017AttentionIA}. We compare each token with each other and do token-level fusion with each other to learn an attention embedding layer. Then, we use the sequence embedding for classification. We add a simple linear layer over the sequence embedding for classification, and softmax over the given choices. Note that we do not freeze any weights along the process, allowing the transformer and perceptron to be fine-tuned during this process. We also allow the knowledge embeddings to be modified through this back-propagation. Hyperparameters are noted in Section \ref{ht}. We also ablate our use of this extra attention layer, showing that it is important to learn comparisons between knowledge embeddings. For BERT baselines, we use the process in \citet{Devlin2018BERTPO}, and use the $[CLS]$ token, without attention, for classification.

\section{Analysis}

\subsection{Hyperparameter Tuning}
\label{ht}

\begin{figure} 
    \centering
  \includegraphics[width=0.48\textwidth]{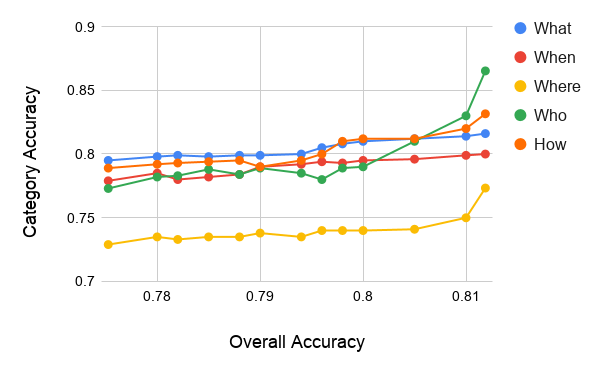}
\caption{Example of B. MOOD accuracy across categories during hyperparameter turning. Values to the right are closer to the maxima.}
\label{fig:graph}
\end{figure}


For hyperparameter tuning with BERT, we find that grid search is the best method. We tune various hyperparameters, including batch size, learning rate, warmup, and epoch count (for hyperparameter details, see appendix). Graph \ref{fig:graph} shows the results of several hyperparameters on BERT with our additional knowledge bases. We find that B. MOOD seems to correct its deficiencies as it gets closer to the maxima. Interestingly, B. MOOD seems to be naturally good ``What" questions, which commonly require commonsense inference. This could be explained by the effect of the commonsense knowledge graphs, showing that is picking up on commonsense attributes. However, for ``Where" questions, which it requires more information from the text, B. MOOD needs to learn and thus experiences a greater gain as the accuracy gets closer to its maxima. 

We also compare to TriAN \cite{Wang2018YuanfudaoAS}, the previous state-of-the-art. Table \label{tableee} shows our results. For the majority of categories, it seems to begin to be 50/50 between TriAN and B. MOOD, with TriAN showing more strength in commonsense categories. However, B. MOOD begins to get large jumps in accuracy in categories that it is beat in (such as ``Who" and ``Where"). For knowledge embeddings, we use a size of 10 for each knowledge graph, combining for a size 30 knowledge graph embedding. We randomly init each embedding, and if there is more than one embedding for token, we pick one at random \cite{Wang2018YuanfudaoAS}. For BERT fine-tuning, we use a maximum sequence length of 450, a train batch size of 32, four epochs, $1e-5$ learning rate, and a 20\% warmup.

\subsection{Results}

\begin{table}[]
\centering
\begin{tabular}{lcc} \hline
\multirow{2}{*}{System} & \multicolumn{2}{c}{Accuracy} \\ 
 & Dev & Test \\ \hline
Human & 97.4 & 98.0 \\
Logistic Baseline & - & 60.8 \\ 
TriAN \cite{Wang2018YuanfudaoAS} & 76.1 & - \\
BERT$_{LARGE}$ & 82.3 & - \\ \hline 
B. MOOD (with ConceptNet) & 83.1 & - \\
B. MOOD (with WebChild) & 82.7 & - \\
B. MOOD (with ATOMIC) & 82.5 & - \\
B. MOOD (w/o final attention) & 82.4 & - \\
B. MOOD (with all KB) & \textbf{83.3} & \multicolumn{1}{c}{\textbf{80.7}} \\\hline
\end{tabular}
\caption{Results with B. MOOD on task dev and test set. ``with all KB" describes results using all ConceptNet, WebChild, and ATOMIC embeddings. ``Human" and ``Regression Baseline" accuracy is from the shared task paper \cite{Ostermann2018MCScriptAN}. TriAN \cite{Wang2018YuanfudaoAS} uses ConceptNet as features.}
\end{table}

\begin{table}[]
\centering
\begin{tabular}{lccc} \hline
\multirow{2}{*}{Category} & \multicolumn{3}{c}{System Accuracy} \\
 & TriAN & BERT & B. MOOD \\ \hline
What & 79.3 & 81.6 & 84.5 \\
When & 69.4 & 80.0 & 81.3 \\
Where & 75.1 & 77.3 & 78.3 \\
Who & 79.4 & 86.5 & 86.6 \\
How & 76.8 & 83.2 & 83.4 \\ \hline
Overall & 76.1 & 82.3 & \textbf{83.3} \\ \hline
\end{tabular}
\caption{Question type comparison between different models on the shared task: previous state-of-the-art TriAN \cite{Wang2018YuanfudaoAS}, BERT$_{LARGE}$, and B. MOOD (with all 3 knowledge bases).}
\label{tableee}
\end{table}

We show our results and give analysis for BIG MOOD. We show that each of the knowledge bases help the accuracy of our model, and our strongest model involves the union of all three knowledge bases. ConceptNet gives the largest increase, likely because there are the most matches between the prompts and ConceptNet, since ConceptNet covers everyday concepts that are relatively more common. WebChild gives a boost also, but not as large as ConceptNet. ATOMIC gives the smallest boost, likely because $1)$ ATOMIC queries are the longest, and thus, least likely to match, and $2)$ there is not as much inferential commonsense present.

We also note that the base B. MOOD accuracy is higher than the base TriAN \cite{Wang2018YuanfudaoAS} accuracy, the previous state of the art. By appending similar knowledge embeddings, we find that we can bring the TriAN accuracy up to 77.8\%, which is more comparable with B. MOOD. This shows that the additional knowledge bases (ATOMIC, WebChild) contribute to the overall accuracy even without the contextual embeddings. However, we find that the knowledge bases combined with TriAN still do not provide an improvement above that of B. MOOD, and thus, the knowledge bases alone are not enough to capture the necessary information. Instead, the knowledge graphs must be used through combination with contextual embeddings for the most effective model. This shows that BERT may lack the complete amount of information needed to understand this dataset. We also show that the attention is needed to understand the knowledge graphs alongside BERT, showing the importance of learning the different knowledge base embeddings within the text. This highlights the fact that using the knowledge base embeddings is helpful, and also comparisons between different sections of text is helpful for reading comprehension tasks.

\section{Conclusion}

We introduce a method of fine-tuning with graphical embeddings alongside contextual embeddings, B. MOOD. Our method uses three different knowledge bases, and introduces ways of improving both learning speed and knowledge embedding effectiveness. First, we preprocess the dataset, showing that both language model preprocessing and knowledge graph preprocessing is important to the final result. Then, we tune our language model on the shared task, stabilizing the hyperparameter search. We create knowledge graph embeddings and concatenate the embeddings via token realignment. Then, we introduce a final layer of attention that learns both contextual and explicit graph embeddings through contextualization. We show the effect of various knowledge bases, and show our accuracy across various question types. Our model gets fifth on the task leaderboard and outperforms BERT across all question types. We hope that this investigation motivates and furthers additional research in combining commonsense knowledge awareness with transformers.

\section*{Acknowledgments}
The author thanks Maxwell Forbes, as well as the anonymous reviewers, for their helpful and insightful feedback. The author also thanks the shared task organizers for evaluation on the test set. JD is supported by NSF Multimodal.

\bibliography{emnlp-ijcnlp-2019}

\begin{thebibliography}{31}
\expandafter\ifx\csname natexlab\endcsname\relax\def\natexlab#1{#1}\fi

\bibitem[{von Ahn et~al.(2006)von Ahn, Kedia, and Blum}]{Ahn2006VerbosityAG}
Luis von Ahn, Mihir Kedia, and Manuel Blum. 2006.
\newblock Verbosity: a game for collecting common-sense facts.
\newblock In \emph{CHI}.

\bibitem[{Auer et~al.(2007)Auer, Bizer, Kobilarov, Lehmann, Cyganiak, and
  Ives}]{Auer2007DBpediaAN}
S{\"o}ren Auer, Christian Bizer, Georgi Kobilarov, Jens Lehmann, Richard
  Cyganiak, and Zachary~G. Ives. 2007.
\newblock Dbpedia: A nucleus for a web of open data.
\newblock In \emph{ISWC/ASWC}.

\bibitem[{Breen(2004)}]{Breen2004JMdictAJ}
Jim Breen. 2004.
\newblock Jmdict: A japanese-multilingual dictionary.

\bibitem[{Chambers and Jurafsky(2008)}]{Chambers2008UnsupervisedLO}
Nathanael Chambers and Daniel Jurafsky. 2008.
\newblock Unsupervised learning of narrative event chains.
\newblock In \emph{ACL}.

\bibitem[{Dai et~al.(2019)Dai, Yang, Yang, Carbonell, Le, and
  Salakhutdinov}]{Dai2019TransformerXLAL}
Zihang Dai, Zhilin Yang, Yiming Yang, Jaime~G. Carbonell, Quoc~V. Le, and
  Ruslan Salakhutdinov. 2019.
\newblock Transformer-xl: Attentive language models beyond a fixed-length
  context.
\newblock \emph{ArXiv}, abs/1901.02860.

\bibitem[{Devlin et~al.(2018)Devlin, Chang, Lee, and
  Toutanova}]{Devlin2018BERTPO}
Jacob Devlin, Ming-Wei Chang, Kenton Lee, and Kristina Toutanova. 2018.
\newblock Bert: Pre-training of deep bidirectional transformers for language
  understanding.
\newblock \emph{ArXiv}, abs/1810.04805.

\bibitem[{Havasi et~al.(2010)Havasi, Speer, Arnold, Lieberman, Alonso, and
  Moeller}]{Havasi2010OpenMC}
Catherine Havasi, R.~Speer, Kenneth~C. Arnold, Henry Lieberman, Jason~B.
  Alonso, and Jesse Moeller. 2010.
\newblock Open mind common sense: Crowd-sourcing for common sense.
\newblock In \emph{Collaboratively-Built Knowledge Sources and AI}.

\bibitem[{Lai et~al.(2017)Lai, Xie, Liu, Yang, and Hovy}]{Lai2017RACELR}
Guokun Lai, Qizhe Xie, Hanxiao Liu, Yiming Yang, and Eduard~H. Hovy. 2017.
\newblock Race: Large-scale reading comprehension dataset from examinations.
\newblock In \emph{EMNLP}.

\bibitem[{Lenat and Guha(1989)}]{Lenat1989BuildingLK}
Douglas~B. Lenat and Ramanathan~V. Guha. 1989.
\newblock Building large knowledge-based systems; representation and inference
  in the cyc project.

\bibitem[{Lin et~al.(2017)Lin, Sun, and Han}]{Lin2017ReasoningWH}
Hongyu Lin, Le~Sun, and Xianpei Han. 2017.
\newblock Reasoning with heterogeneous knowledge for commonsense machine
  comprehension.
\newblock In \emph{EMNLP}.

\bibitem[{Liu et~al.(2019)Liu, Ott, Goyal, Du, Joshi, Chen, Levy, Lewis,
  Zettlemoyer, and Stoyanov}]{Liu2019RoBERTaAR}
Yinhan Liu, Myle Ott, Naman Goyal, Jingfei Du, Mandar~S. Joshi, Danqi Chen,
  Omer Levy, Miranda Paige~Linscott Lewis, Luke~S. Zettlemoyer, and Veselin
  Stoyanov. 2019.
\newblock Roberta: A robustly optimized bert pretraining approach.
\newblock \emph{ArXiv}, abs/1907.11692.

\bibitem[{Merkhofer et~al.(2018)Merkhofer, Henderson, Bloom, Strickhart, and
  Zarrella}]{Merkhofer2018MITREAS}
Elizabeth~M. Merkhofer, John~C. Henderson, David Bloom, Laura Strickhart, and
  Guido Zarrella. 2018.
\newblock Mitre at semeval-2018 task 11: Commonsense reasoning without
  commonsense knowledge.
\newblock In \emph{SemEval@NAACL-HLT}.

\bibitem[{Meyer and Gurevych(2012)}]{Meyer2012WiktionaryAN}
Christian~M. Meyer and Iryna Gurevych. 2012.
\newblock Wiktionary: A new rival for expert-built lexicons? exploring the
  possibilities of collaborative lexicography.

\bibitem[{Miller(1992)}]{Miller1992WORDNETAL}
George~A. Miller. 1992.
\newblock Wordnet: A lexical database for english.
\newblock \emph{Commun. ACM}, 38:39--41.

\bibitem[{Ostermann et~al.(2018)Ostermann, Modi, Roth, Thater, and
  Pinkal}]{Ostermann2018MCScriptAN}
Simon Ostermann, Ashutosh Modi, Michael Roth, Stefan Thater, and Manfred
  Pinkal. 2018.
\newblock Mcscript: A novel dataset for assessing machine comprehension using
  script knowledge.
\newblock \emph{ArXiv}, abs/1803.05223.

\bibitem[{Peters et~al.(2018)Peters, Neumann, Iyyer, Gardner, Clark, Lee, and
  Zettlemoyer}]{Peters2018DeepCW}
Matthew~E. Peters, Mark Neumann, Mohit Iyyer, Matt Gardner, Christopher Clark,
  Kenton Lee, and Luke~S. Zettlemoyer. 2018.
\newblock Deep contextualized word representations.
\newblock \emph{ArXiv}, abs/1802.05365.

\bibitem[{Porter(1980)}]{Porter1980AnAF}
Martin~F. Porter. 1980.
\newblock An algorithm for suffix stripping.
\newblock \emph{Program}, 40:211--218.

\bibitem[{Sap et~al.(2018)Sap, Bras, Allaway, Bhagavatula, Lourie, Rashkin,
  Roof, Smith, and Choi}]{Sap2018ATOMICAA}
Maarten Sap, Ronan~Le Bras, Emily Allaway, Chandra Bhagavatula, Nicholas
  Lourie, Hannah Rashkin, Brendan Roof, Noah~A. Smith, and Yejin Choi. 2018.
\newblock Atomic: An atlas of machine commonsense for if-then reasoning.
\newblock \emph{ArXiv}, abs/1811.00146.

\bibitem[{Schwartz et~al.(2019)Schwartz, Dodge, Smith, and
  Etzioni}]{Schwartz2019GreenA}
Roy Schwartz, Jesse Dodge, Noah~A. Smith, and Oren Etzioni. 2019.
\newblock Green ai.

\bibitem[{Speer and Havasi(2013)}]{Speer2013ConceptNet5A}
R.~Speer and Catherine Havasi. 2013.
\newblock Conceptnet 5: A large semantic network for relational knowledge.
\newblock In \emph{The People's Web Meets NLP}.

\bibitem[{Strubell et~al.(2019)Strubell, Ganesh, and
  McCallum}]{Strubell2019EnergyAP}
Emma Strubell, Ananya Ganesh, and Andrew McCallum. 2019.
\newblock Energy and policy considerations for deep learning in nlp.
\newblock \emph{ArXiv}, abs/1906.02243.

\bibitem[{Sun et~al.(2019)Sun, Wang, Li, Feng, Chen, Zhang, Tian, Zhu, Tian,
  and Wu}]{Sun2019ERNIEER}
Yu~Sun, Shuohuan Wang, Yukun Li, Shikun Feng, Xuyi Chen, Han Zhang, Xin Tian,
  Danxiang Zhu, Hao Tian, and Hua Wu. 2019.
\newblock Ernie: Enhanced representation through knowledge integration.
\newblock \emph{ArXiv}, abs/1904.09223.

\bibitem[{Tandon et~al.(2017)Tandon, de~Melo, and
  Weikum}]{tandon-etal-2017-webchild}
Niket Tandon, Gerard de~Melo, and Gerhard Weikum. 2017.
\newblock \href {https://www.aclweb.org/anthology/P17-4020} {{W}eb{C}hild 2.0 :
  Fine-grained commonsense knowledge distillation}.
\newblock In \emph{Proceedings of {ACL} 2017, System Demonstrations}, pages
  115--120, Vancouver, Canada. Association for Computational Linguistics.

\bibitem[{Vaswani et~al.(2017)Vaswani, Shazeer, Parmar, Uszkoreit, Jones,
  Gomez, Kaiser, and Polosukhin}]{Vaswani2017AttentionIA}
Ashish Vaswani, Noam Shazeer, Niki Parmar, Jakob Uszkoreit, Lawrence Jones,
  Aidan~N. Gomez, Lukasz Kaiser, and Illia Polosukhin. 2017.
\newblock Attention is all you need.
\newblock In \emph{NIPS}.

\bibitem[{Wang(2018)}]{Wang2018YuanfudaoAS}
Liang Wang. 2018.
\newblock Yuanfudao at semeval-2018 task 11: Three-way attention and relational
  knowledge for commonsense machine comprehension.
\newblock In \emph{SemEval@NAACL-HLT}.

\bibitem[{Wu et~al.(2016)Wu, Schuster, Chen, Le, Norouzi, Macherey, Krikun,
  Cao, Gao, Macherey, Klingner, Shah, Johnson, Liu, Kaiser, Gouws, Kato, Kudo,
  Kazawa, Stevens, Kurian, Patil, Wang, Young, Smith, Riesa, Rudnick, Vinyals,
  Corrado, Hughes, and Dean}]{Wu2016GooglesNM}
Yonghui Wu, Mike Schuster, Zhifeng Chen, Quoc~V. Le, Mohammad Norouzi, Wolfgang
  Macherey, Maxim Krikun, Yuan Cao, Qin Gao, Klaus Macherey, Jeff Klingner,
  Apurva Shah, Melvin Johnson, Xiaobing Liu, Lukasz Kaiser, Stephan Gouws,
  Yoshikiyo Kato, Taku Kudo, Hideto Kazawa, Keith Stevens, George Kurian,
  Nishant Patil, Wei Wang, Cliff Young, Jason Smith, Jason Riesa, Alex Rudnick,
  Oriol Vinyals, Gregory~S. Corrado, Macduff Hughes, and Jeffrey Dean. 2016.
\newblock Google's neural machine translation system: Bridging the gap between
  human and machine translation.
\newblock \emph{ArXiv}, abs/1609.08144.

\bibitem[{Yang and Mitchell(2017)}]{Yang2017LeveragingKB}
Bishan Yang and Tom~M. Mitchell. 2017.
\newblock Leveraging knowledge bases in lstms for improving machine reading.
\newblock In \emph{ACL}.

\bibitem[{Yang et~al.(2019)Yang, Dai, Yang, Carbonell, Salakhutdinov, and
  Le}]{Yang2019XLNetGA}
Zhilin Yang, Zihang Dai, Yiming Yang, Jaime~G. Carbonell, Ruslan Salakhutdinov,
  and Quoc~V. Le. 2019.
\newblock Xlnet: Generalized autoregressive pretraining for language
  understanding.
\newblock \emph{ArXiv}, abs/1906.08237.

\bibitem[{Zellers et~al.(2018)Zellers, Bisk, Schwartz, and
  Choi}]{Zellers2018SWAGAL}
Rowan Zellers, Yonatan Bisk, Roy Schwartz, and Yejin Choi. 2018.
\newblock Swag: A large-scale adversarial dataset for grounded commonsense
  inference.
\newblock \emph{ArXiv}, abs/1808.05326.

\bibitem[{Zhang et~al.(2017)Zhang, Liu, Luan, Xu, and Sun}]{Zhang2017PriorKI}
Jiacheng Zhang, Yang Liu, Huanbo Luan, Jingfang Xu, and Maosong Sun. 2017.
\newblock Prior knowledge integration for neural machine translation using
  posterior regularization.
\newblock \emph{ArXiv}, abs/1811.01100.

\bibitem[{Zhu et~al.(2015)Zhu, Kiros, Zemel, Salakhutdinov, Urtasun, Torralba,
  and Fidler}]{Zhu2015AligningBA}
Yukun Zhu, Jamie~Ryan Kiros, Richard~S. Zemel, Ruslan Salakhutdinov, Raquel
  Urtasun, Antonio Torralba, and Sanja Fidler. 2015.
\newblock Aligning books and movies: Towards story-like visual explanations by
  watching movies and reading books.
\newblock \emph{2015 IEEE International Conference on Computer Vision (ICCV)},
  pages 19--27.

\end{thebibliography}
\bibliographystyle{acl_natbib}

\newpage
\appendix
\section{Appendices}
\subsection{Hyperparameters}
Seen in Table \ref{er-hyp} is a list of hyperparameters for our experiments. We use the same parameters for both uses of explicit knowledge embeddings. 

\begin{table}[h]
\small
\centering
\begin{tabular}{ |l c|}
\hline
\multicolumn{2}{|c|}{Explicit Knowledge Embeddings}\\
Embedding size & 10 \\
Knowledge bases used & 3 \\
\multicolumn{2}{|c|}{BERT Fine-Tuning}\\
Maximum sequence length & 450 \\
Train batch size & 32 \\
Learning rate & 1e-5 \\
Epochs & 4 \\
Warmup & 20\% \\
\multicolumn{2}{|c|}{TriAN Parameters}\\
Optimizer & adamax \\
Learning rate & 2e-3 \\
Batch size & 32 \\
Hidden size & 96 \\
RNN type & lstm \\
Embedding dropout & 0.4 \\
\hline
\end{tabular}
\caption{Hyperparameters used throughout experiments. TriAN parameters are used for TriAN comparison only.}
\label{er-hyp}
\end{table}

\end{document}